\documentclass[11pt]{article}

\usepackage[final]{acl}

\usepackage{latexsym}

\usepackage{xcolor}

\usepackage[utf8]{inputenc}

\usepackage{microtype}

\usepackage[english,bidi=default]{babel}
\babelfont{rm}{TeXGyreTermesX}

\usepackage{inconsolata}

\usepackage{hyperref}

\usepackage{pdfpages}
\usepackage{graphicx}   %
\usepackage{subcaption} %
\usepackage{booktabs,multirow}
\usepackage{adjustbox}
\usepackage{xeCJK}
\usepackage{xcolor}
\usepackage{xeCJKfntef}

\usepackage{soul}
\sethlcolor{yellow}
\usepackage{amsmath}
\usepackage{bm}
\usepackage{makecell}
\usepackage{xspace}
\usepackage{xcolor}
\usepackage{cleveref}

\crefname{figure}{Fig.}{Figs.}
\Crefname{figure}{Fig.}{Figs.}%

\newcommand{\datasetname}{\textcolor{black}{\textsc{DimStance}}\xspace}

\usepackage{fontspec}
\usepackage{xeCJK}
\title{DimStance: Multilingual Datasets for Dimensional Stance Analysis}

\author{
\textbf{Jonas Becker}$^{1,*}$, \textbf{Liang-Chih Yu}$^{2,*}$, \textbf{Shamsuddeen Hassan Muhammad}$^{3}$, \textbf{Jan Philip Wahle}$^{1}$, \\
\textbf{Terry Ruas}$^{1}$, \textbf{Idris Abdulmumin}$^{4}$, \textbf{Lung-Hao Lee}$^{5}$, \textbf{Nelson Odhiambo}$^{6}$, \textbf{Lilian Wanzare}$^{6}$,\\
\textbf{Wen-Ni Liu}$^{5}$, \textbf{Tzu-Mi Lin}$^{5}$, \textbf{Zhe-Yu Xu}$^{5}$, \textbf{Ying-Lung Lin}$^{7}$, \textbf{Jin Wang}$^{8}$, \\
\textbf{Maryam Ibrahim Mukhtar}$^{9}$, \textbf{Bela Gipp}$^{1}$, \textbf{Saif M. Mohammad}$^{10}$ \vspace{+0.5em}
\\
$^{1}$University of G\"ottingen, $^{2}$Yuan Ze University, $^{3}$Imperial College London, $^{4}$University of Pretoria, \\
$^{5}$National Yang Ming Chiao Tung University, $^{6}$Maseno University, $^{7}$Central Police University Taiwan, \\
$^{8}$Yunnan University, $^{9}$Bayero University Kano, $^{10}$National Research Council Canada \vspace{+0.5em}
\\
\textbf{*Contact:} \texttt{jonas.becker@uni-goettingen.de, lcyu@saturn.yzu.edu.tw}
}

\begin{document}
\maketitle

\begin{center}
\vspace{-1em}
\fbox{\parbox{0.95\linewidth}{\centering\small\bfseries
Offensive Content Warning: This paper contains examples of offensive language for research purposes.
}}
\vspace{0.5em}
\end{center}

\begin{abstract}
Stance detection is an established task that classifies an author's attitude toward a specific target into categories such as Favor, Neutral, and Against. Beyond categorical stance labels, we leverage a long-established affective science framework to model stance along real-valued dimensions of valence (negative–positive) and arousal (calm–active). This dimensional approach captures nuanced affective states underlying stance expressions, enabling fine-grained stance analysis. To this end, we introduce \datasetname, the first dimensional stance resource with valence–arousal (VA) annotations. This resource comprises 11,746 target aspects in 7,365 texts across five languages (English, German, Chinese, Nigerian Pidgin, and Swahili) and two domains (politics and environmental protection). To facilitate the evaluation of stance VA prediction, we formulate the dimensional stance regression task, analyze cross-lingual VA patterns, and benchmark pretrained and large language models under regression and prompting settings. Results show competitive performance of fine-tuned LLM regressors, persistent challenges in low-resource languages, and limitations of token-based generation. \datasetname provides a foundation for multilingual, emotion-aware, stance analysis and benchmarking.
\end{abstract}

\section{Introduction}

Stance detection aims to automatically identify an author's stance from text toward a specific target (e.g., a person, organization, or policy) \citep{Chandra:81, MohammadKSZ16a, 10.1145/3369026, aldayel2021stance}. The stance is typically classified into predefined labels such as {\texttt{Favor}}, {\texttt{Against}}, or {\texttt{Neither (Neutral)}}. For example, given the sentence ``\textit{We must act now to protect the environment}.'' and the target \textit{environment}, a stance detection system should label the stance as {\texttt{Favor}}. In recent years, a wide range of applications and datasets for stance detection have been proposed \citep{hardalov-etal-2022-survey, Zhang2024SurveyStanceDetection, pangtey2025large}, alongside the emergence of specialized task settings such as multimodal \cite{zhou-etal-2025-media-frames-stance, zhang-etal-2025-mad} and conversational \cite{ding-etal-2025-zero, marreddy-etal-2025-usdc} stance detection.

\begin{figure*}[tbp]
    \centering    \includegraphics[width=0.9\linewidth]{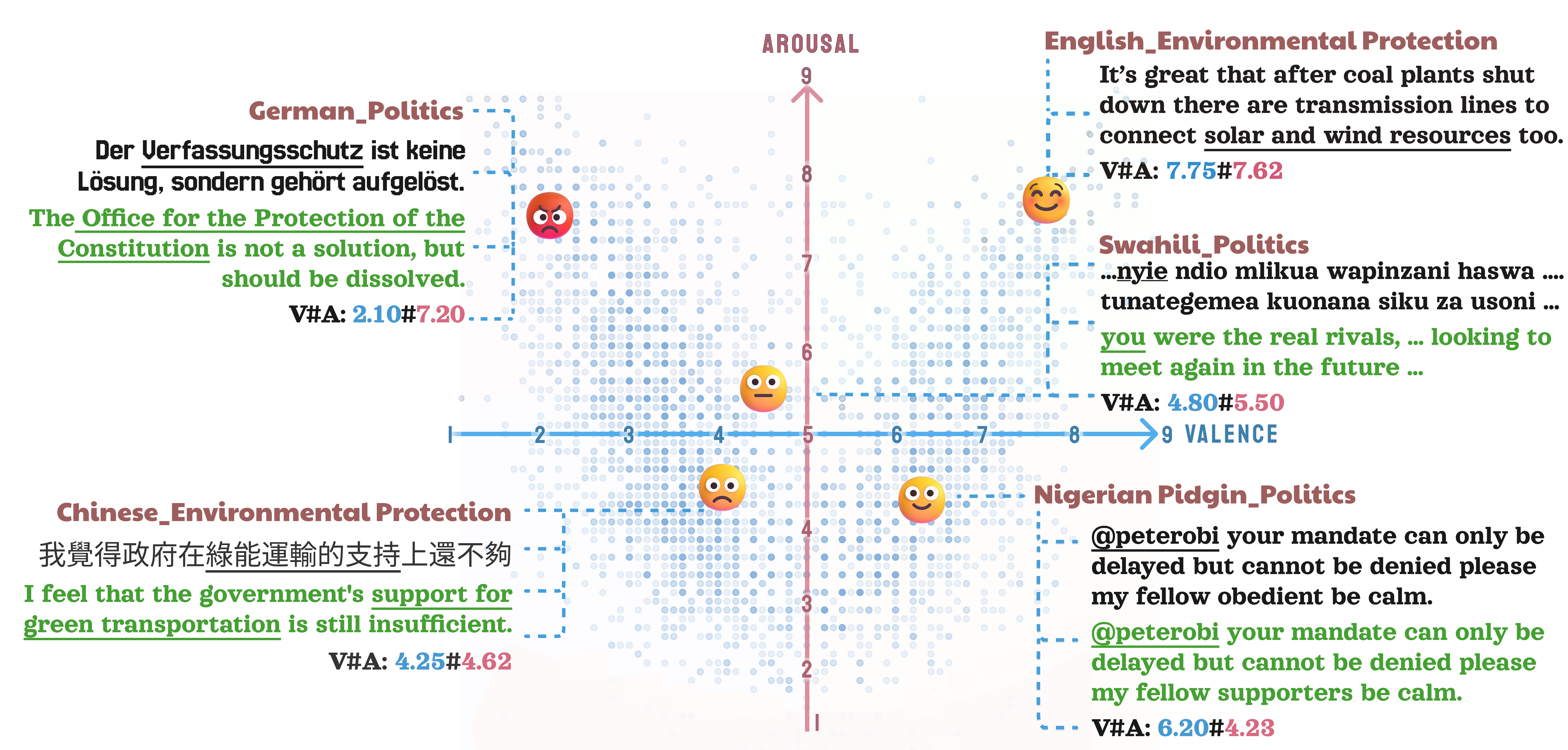}
    \caption{\textbf{Valence-Arousal space.} Illustrative examples of multilingual stance expressions. The blue dots are a scatter of valence-arousal scores, showing a U-shaped distribution. Five examples of the \datasetname datasets are described as \textit{Language\_Domain} with their corresponding translation printed in green. Underlined text indicates the target aspect. V\#A reports the Valence (V) and Arousal (A) annotation for the target aspect.} 
    \label{fig:VA scatter}
\end{figure*}

Despite these advances, existing stance detection work still relies on categorical labels for classification.
In many settings, stances are rarely purely ``for" or ``against". Even expressions sharing the same stance can exhibit substantial variation in affective tone and intensity, motivating the need to identify the affective states underlying stance expressions.
By measuring the emotions that drive stance at scale, we can sharpen our understanding of why opinions harden—and, in turn, inform strategies to mitigate the intense polarization seen in debates such as climate change or politics.

To address these limitations, we draw insights from long-established affective science frameworks \citep{russell1980circumplex,russell2003core}, which model evaluative attitudes along dimensions of \textbf{valence} (displeasure--pleasure/negative--positive) and \textbf{arousal} (sluggish--excited/calm--active). 
As illustrated by the example stance expressions in  \Cref{fig:VA scatter}, English and Nigerian Pidgin would both receive {\texttt{Favor}} in a categorical scheme, yet the dimensional approach further distinguishes the former as strong, highly aroused support and the latter as mild, low-arousal approval.
Likewise, both German and Chinese correspond to {\texttt{Against}}, although the former reflects emotionally charged opposition while the latter conveys calm disagreement.
This enhanced granularity captures the nuances of evaluative attitudes, facilitating a finer-grained stance analysis across languages and domains.
Thus, this enriched form of stance detection can be of considerable value in analyzing and understanding public opinion on complex issues such as climate change, vaccine hesitancy, the role of AI in society, and government policies.

To our knowledge, no existing stance detection resources adopt a valence-arousal (VA) representation. To fill this gap, we introduce the \datasetname, the first dimensional stance resource manually annotated with continuous VA scores. The datasets contain 11,746 target aspects across 7,365 texts and cover five high- and low-resource languages (\textit{Chinese, English, German, Nigerian Pidgin}, and \textit{Swahili}) across two domains (\textit{politics} and \textit{environmental protection}). We introduced a new task, termed dimensional stance regression, which requires systems to predict continuous VA scores for each given target aspect within a stance expression.  

We evaluate a range of pre-trained language models (PLMs) and large language models (LLMs) under regression and prompting settings and analyze cross-lingual VA patterns. We observe that fine-tuned LLM regressors generally outperform prompted LLMs and PLM regressors. Yet, prompting-based LLMs can serve as a cost-efficient alternative. The results further highlight challenges in low-resource languages and limitations of token-based VA prediction. \textbf{We make our datasets public}\footnote{\url{https://github.com/DimABSA/DimABSA2026}}, providing a pivotal foundation for multilingual dimensional stance analysis to address language-specific nuances and challenges.
Our contributions are summarized as follows: 
\begin{itemize}
    \item We present the first \datasetname~datasets, spanning five high- and low-resource languages and two domains.
    \item We introduce a dimensional stance regression task that integrates valence–arousal representations into the stance detection framework.  
    \item  We provide baseline results across PLM regressors, prompted LLMs, and fine-tuned LLM regressors, facilitating future benchmarking for dimensional stance analysis.
\end{itemize}
\noindent All of the data and baselines will be made freely available on the project website.

\begin{table*}[t]
\centering
\small
\setlength{\tabcolsep}{4pt}
\renewcommand{\arraystretch}{1.5}
\begin{adjustbox}{max width=\linewidth}
\begin{tabular}{l l l c c c c c c}
\hline
\textbf{Language} & \textbf{Domain} & \textbf{Source(s)} & \textbf{\#Ann.} & 
\textbf{Train} & \textbf{Dev} & \textbf{Test} & \textbf{Total} & \textbf{RMSE} (V\#A) \\
\hline
English & Environmental Protection& EZ-STANCE; Reddit& 7 
& 922 / 2059& 200 / 339& 1020 / 1813& 2142 / 4211& 1.696\#2.195\\

German & Politics& Wahl-O-Mat Archive & 8 
& 683 / 1335& 34 / 75& 263 / 438& 980 / 1848& 1.056\#1.818 \\

Chinese & Environmental Protection & Threads Platform & 9 
& 683 / 1091& 34 / 49& 263 / 898& 980 / 2038& 0.930\#0.902 \\

Nigerian Pidgin & Politics& X Platform & 5 
& 1049 / 1118& 119 / 122& 331 / 343& 1499 / 1583& 2.125\#2.881 \\

Swahili & Politics& X Platform& 5 
& 1375 / 1622& 123 / 145& 266 / 299& 1764 / 2066& 2.958\#2.892 \\

\hline
\end{tabular}
\end{adjustbox}
\caption{\textbf{Dataset overview.} For each language, we report the domain, data source(s), total number of annotators, train/dev/test splits, and aggregated totals. All counts are reported as \textit{number of texts} / \textit{number of target aspects}. Annotation reliability is measured by root mean squared error (RMSE) for valence\#arousal (V\#A) ratings.}
\label{tab:final_data_stats}
\end{table*}

\section{The \datasetname Dataset Collection}

We detail the construction of the dataset, covering data sources, preprocessing steps, and the annotation workflow.

\subsection{Data Sources}

We collect data from five languages, chosen to provide a balance between high-resource languages (English, German, and Chinese) and low-resource languages (Nigerian Pidgin and Swahili). We focus on the domains of politics and environmental protection because they contain a mixture of intense and calm emotional expressions, as well as diverse stances.

\paragraph{English.} Data for the training split is collected from the environmental protection domain of EZ-STANCE \citep{zhao-caragea-2024-ez}. The dev and test splits are obtained from Reddit texts\footnote{\url{https://reddit.com/}. Version: 2025-07-01}, using the same keywords as in EZ-STANCE.

\paragraph{German.} Sampled from Wahl-O-Mat Archive, provided by the Federal Agency for Civic Education of Germany\footnote{\url{https://www.bpb.de/themen/wahl-o-mat/556865/datensaetze-des-wahl-o-mat/}. Version: 2026-03-25.}. The data contains responses by political parties to political statements.

\paragraph{Chinese.} Collected messages from the Threads platform\footnote{\url{https://www.threads.com/}. Version: 2025-10-15}. Crawling is performed using a predefined set of Chinese query keywords about environmental protection. 

\paragraph{Nigerian Pidgin.} Sampled posts and comments from the X platform\footnote{\url{https://x.com/}. Version: 2023-12-31}. The discussions concern Nigerian elections, i.e., politics, ranging from January 1st to March 8th, 2023.

\paragraph{Swahili.} Combined data from Afrisenti \citep{muhammad-etal-2023-afrisenti}, HateSpeech\_Kenya \citep{Ombui}, and Politikweli \citep{Amol_PolitikWeli}, covering political tweets from the X platform.

\begin{figure*}[ht]
    \centering

    \begin{subfigure}[b]{0.32\textwidth}
        \centering
        \includegraphics[width=\textwidth]{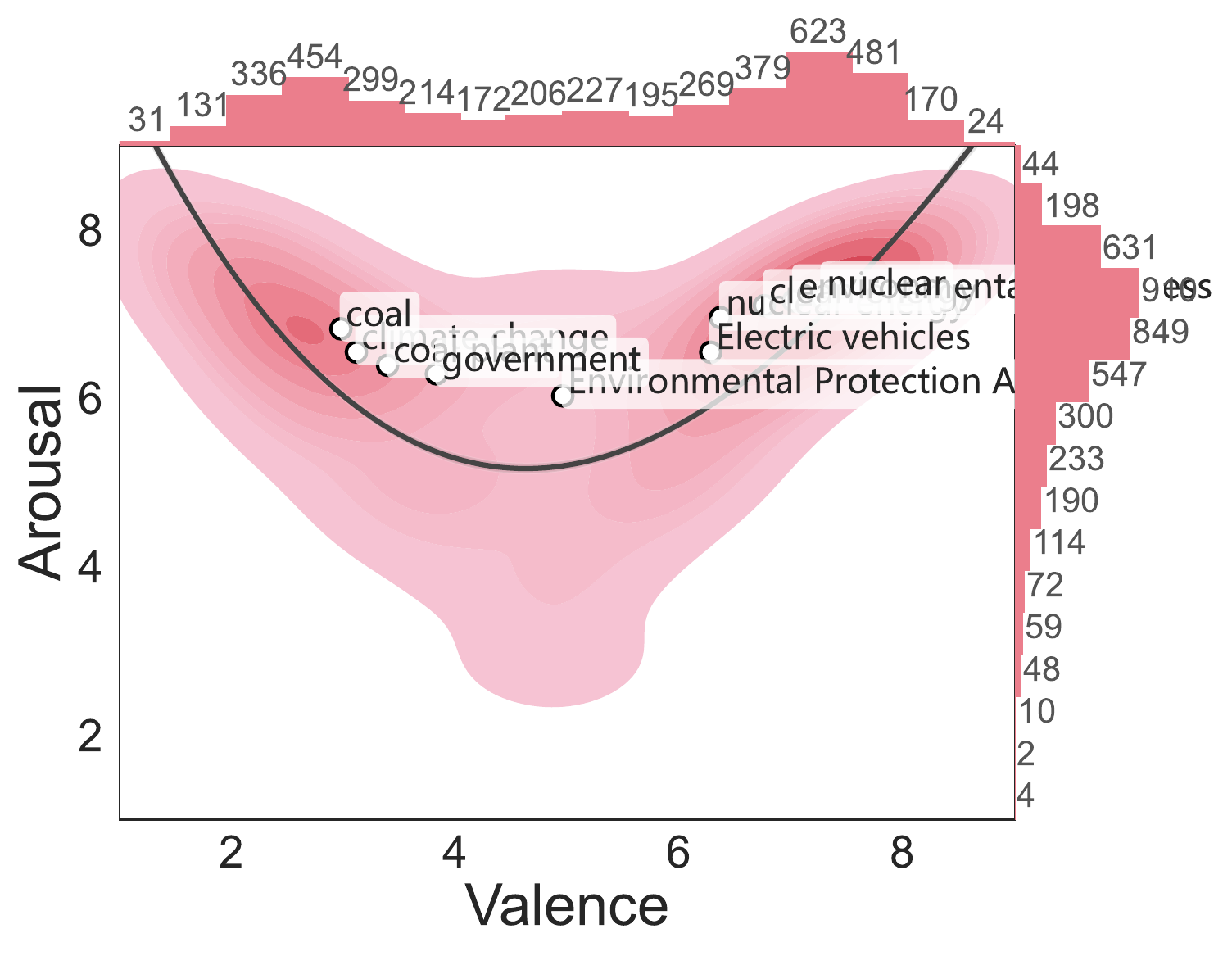}
        \caption{English (env. protection)}
        \label{fig:sub1}
    \end{subfigure}
    \hfill
    \begin{subfigure}[b]{0.32\textwidth}
        \centering
        \includegraphics[width=\textwidth]{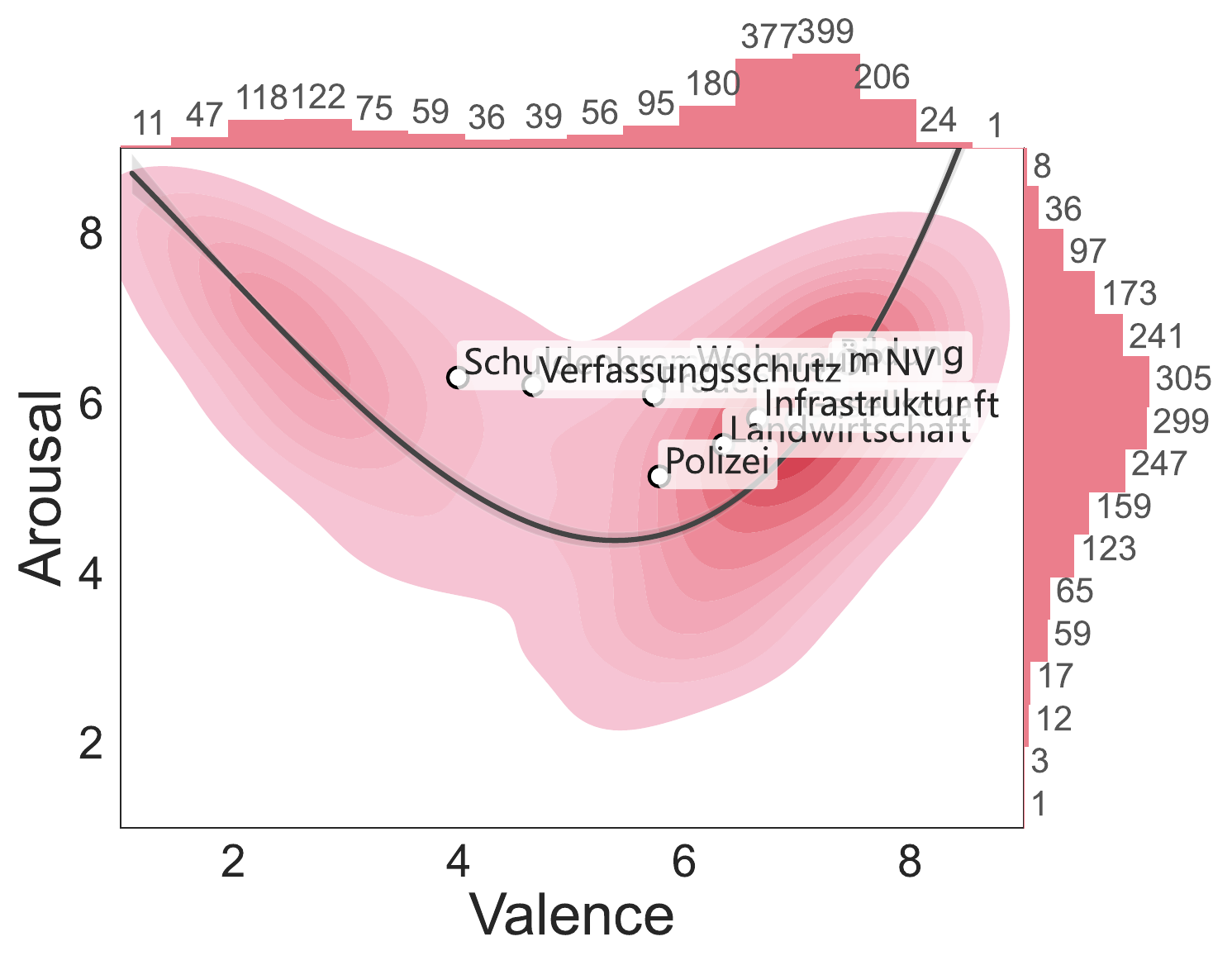}
        \caption{German (politics)}
        \label{fig:sub2}
    \end{subfigure}
    \hfill
    \begin{subfigure}[b]{0.32\textwidth}
        \centering
        \includegraphics[width=\textwidth]{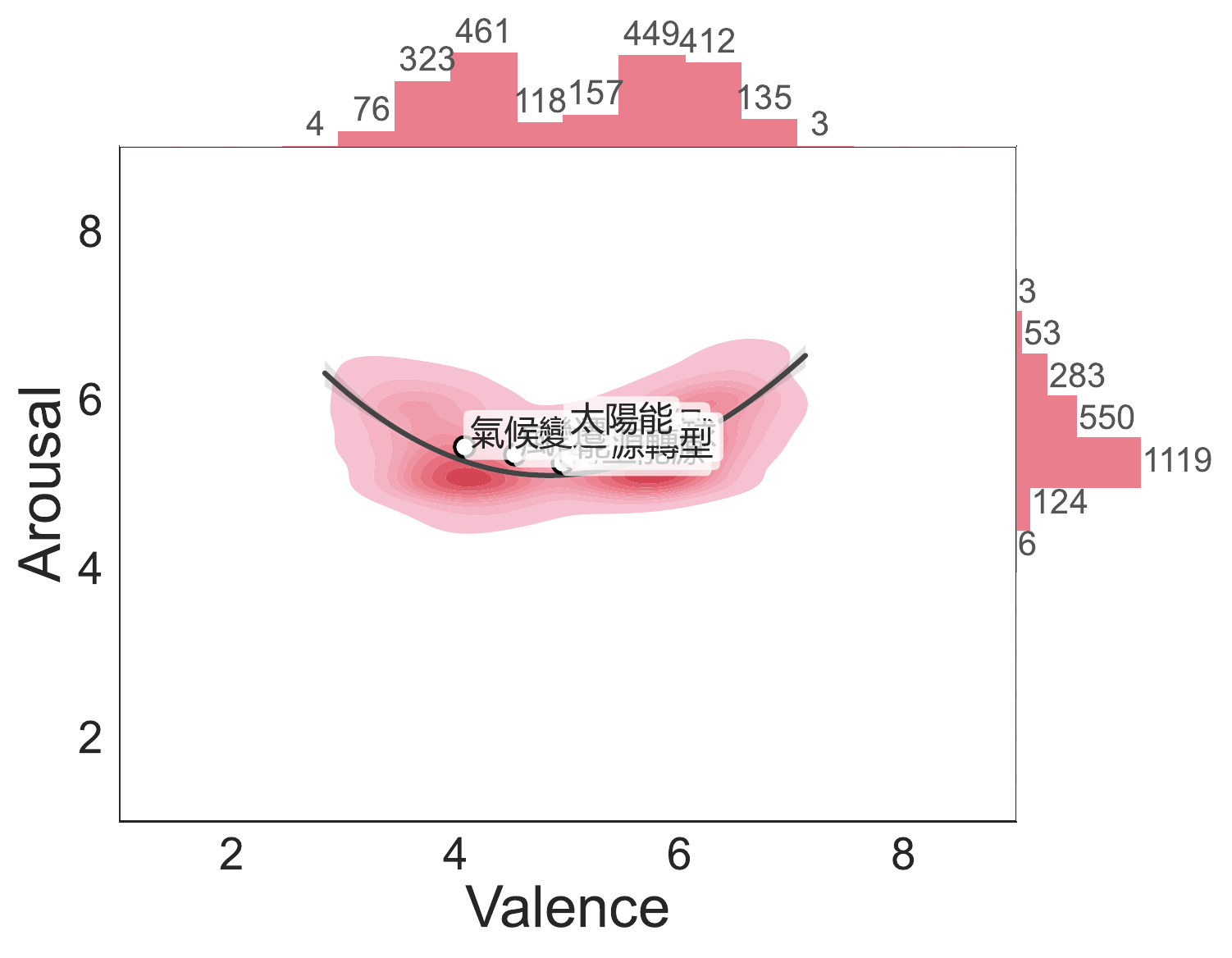}
        \caption{Chinese (env. protection)}
        \label{fig:sub3}
    \end{subfigure}

    \vspace{0.6em}

    \begin{subfigure}[b]{0.32\textwidth}
        \centering
        \includegraphics[width=\textwidth]{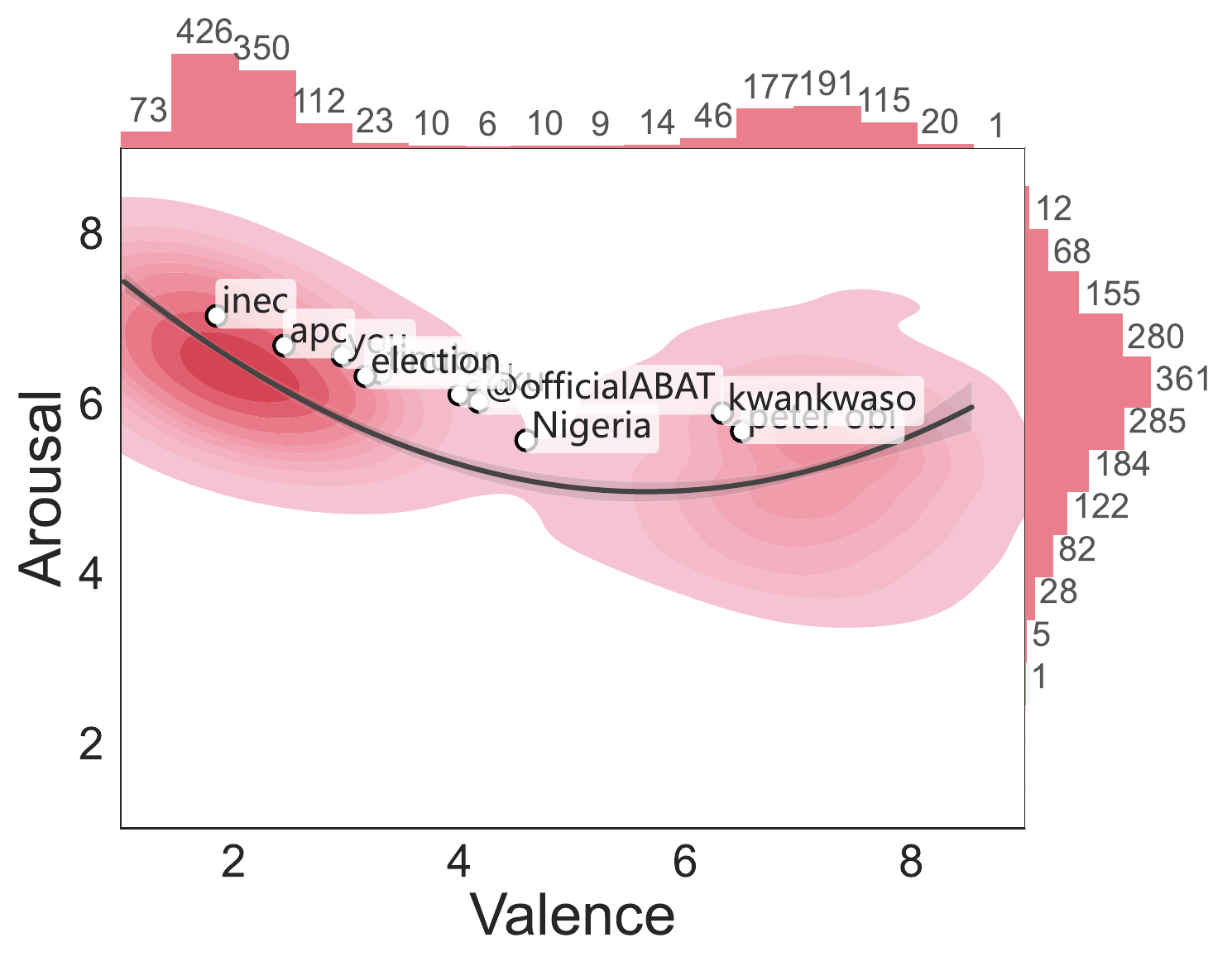}
        \caption{Nigerian Pidgin (politics)}
        \label{fig:sub4}
    \end{subfigure}
    \begin{subfigure}[b]{0.32\textwidth}
        \centering
        \includegraphics[width=\textwidth]{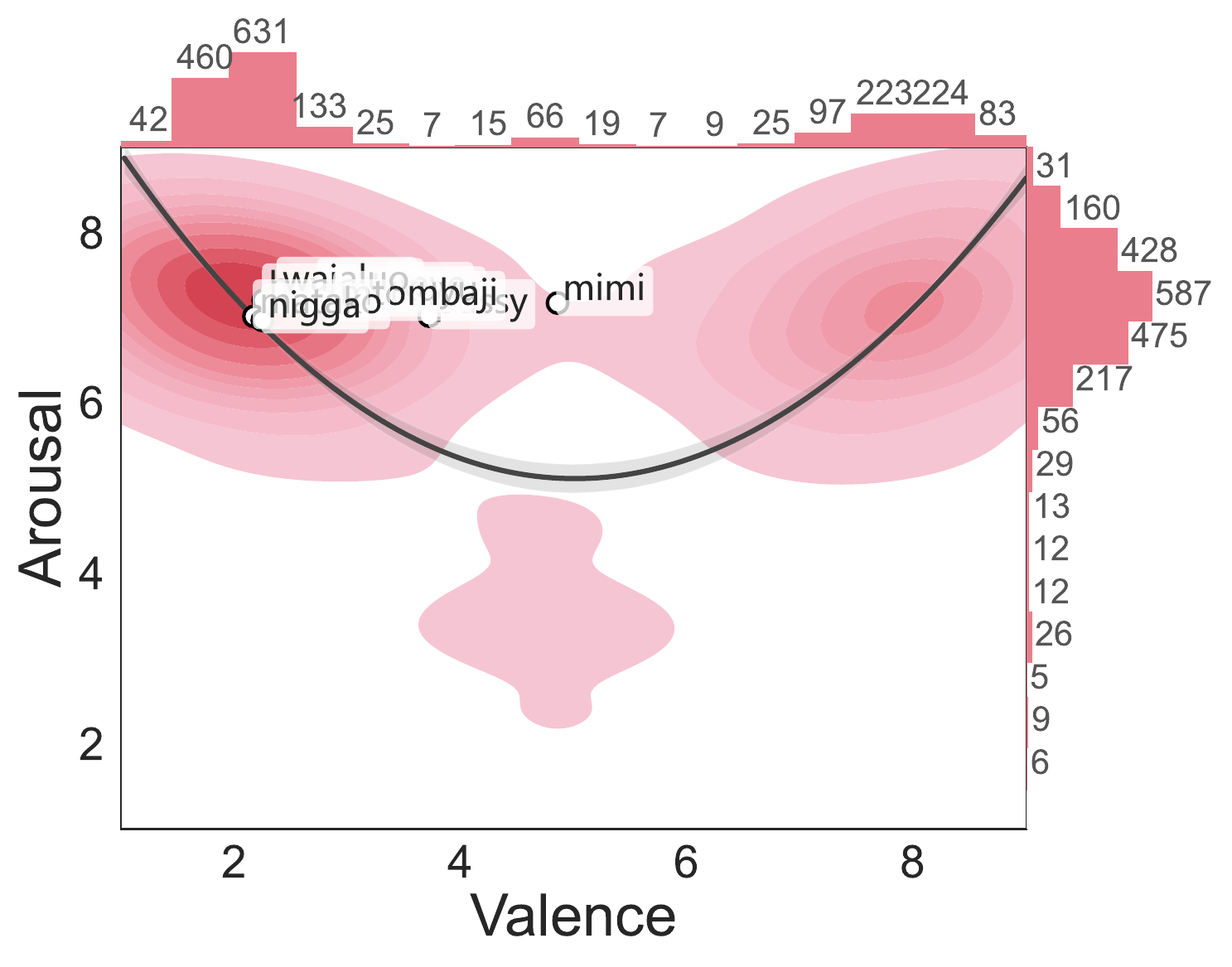}
        \caption{Swahili (politics)}
        \label{fig:sub5}
    \end{subfigure}

    \caption{\textbf{VA distributions by language.} For each language, we show a kernel density plot of the VA scatter space. We show the top ten most prevalent targets in the dataset. An extensive list of the top targets is included in \Cref{tab:top10aspects}.}
    \label{fig:va_scatter}
\end{figure*}

\subsection{Pre-processing and Quality Control}

We preprocess all data by removing duplicates, invisible characters, and faulty encoding.
For raw texts that do not provide marked target aspects already, we use LLMs (eng: \texttt{ChatGPT-4o}, deu: \texttt{meta-llama/Llama-3.3-70B-Instruct}, pcm/swa: \texttt{Gemini-2.0-flash-lite}) to extract candidate target aspects and use regex text matching to identify their exact token positions.

\subsection{Annotation Process}

To ensure the validity of automatically extracted target aspects, we conduct a majority vote with five human annotators (cf. annotation guidelines in \Cref{app:annotation_guidelines}).
If a target aspect is marked as invalid by the majority, it is discarded.
Each valid target aspect is annotated by five native speakers of the respective language, paid at least minimum wage. 
Since a single text may contain multiple target aspects, we present each target-text pair as an independent instance for annotation, rating each on the valence and arousal scales.
The valence scale ranges from 1 \textit{(highly negative)} to 5 \textit{(neutral)} to 9 \textit{(highly positive)}.
The arousal scale ranges from 1 \textit{(highly unaroused)} to 9 \textit{(highly aroused)}.
We average the ratings of the five annotators to obtain the final gold valence and arousal values.
This setup reduces annotator-specific noise, yielding more stable and consistent valence–arousal labels.
We report the annotation guidelines (cf. \Cref{app:annotation_guidelines}) and the annotation interface (cf. \Cref{app:annotation_interface}).  

\subsection{Annotators' Reliability}

The annotation reliability is measured using the Root Mean Squared Error (RMSE) for both valence and arousal ratings.
For each annotator, the RMSE is calculated by averaging the squared differences between their scores and the gold standard (defined as the mean score of all five annotators) across all instances. 
The final reliability is reported as the mean RMSE across all five annotators, as shown in \Cref{tab:final_data_stats}.  

In general, arousal annotations show lower reliability than valence annotations, suggesting that arousal is more challenging for humans to evaluate. 
This is likely because valence provides more distinct perceptual cues, whereas arousal is often subtle and context-dependent. 
This observation is consistent with previous studies reporting lower reliability for arousal annotations \cite{buechel-hahn-2017-emobank, mohammad2018obtaining, lee2022chinese}.

\subsection{Final Data Statistics}

\Cref{tab:final_data_stats} summarizes the dataset across five languages, including data sources, number of annotators, and train/dev/test splits. Each split is shown as texts/target aspects. The dataset comprises a total of 7,365 texts and 11,746 target aspects across all languages.

\newcommand{\msd}{\textsubscript{\text{mean}}\,{\tiny(SD)}}

\begin{table*}[t]
\centering
\small
\setlength{\tabcolsep}{4pt}
\renewcommand{\arraystretch}{1.25}
\begin{adjustbox}{max width=\linewidth}
\begin{tabular}{lccc ccc ccc}
\toprule
& \multicolumn{3}{c}{Positive ($V>5.5$)}
& \multicolumn{3}{c}{Neutral ($4.5<=V<=5.5$)}
& \multicolumn{3}{c}{Negative ($V<4.5$)} \\
\cmidrule(lr){2-4} \cmidrule(lr){5-7} \cmidrule(lr){8-10}
Language
& \% (n) & V\msd & A\msd
& \% (n) & V\msd & A\msd
& \% (n) & V\msd & A\msd \\
\midrule

English
& 49.8\% {\tiny(2097)} & 7.053 {\tiny(0.679)} & 6.845 {\tiny(0.872)}
& 11.3\% {\tiny(477)}  & 4.979 {\tiny(0.305)} & 4.870 {\tiny(1.560)}
& 38.9\% {\tiny(1637)} & 2.841 {\tiny(0.747)} & 6.476 {\tiny(1.082)} \\

German
& 69.2\% {\tiny(1278)} & 6.913 {\tiny(0.575)} & 5.589 {\tiny(1.117)}
& 5.4\%  {\tiny(100)}  & 5.036 {\tiny(0.326)} & 4.486 {\tiny(1.451)}
& 25.4\% {\tiny(470)}  & 2.770 {\tiny(0.742)} & 6.494 {\tiny(1.125)} \\

Chinese
& 42.3\% {\tiny(905)}  & 6.068 {\tiny(0.314)} & 5.532 {\tiny(0.435)}
& 17.3\% {\tiny(369)}  & 5.097 {\tiny(0.365)} & 5.236 {\tiny(0.358)}
& 40.4\% {\tiny(864)}  & 3.893 {\tiny(0.329)} & 5.383 {\tiny(0.453)} \\

Nigerian Pidgin
& 35.6\% {\tiny(564)}  & 7.101 {\tiny(0.527)} & 5.229 {\tiny(0.818)}
& 1.2\%  {\tiny(19)}   & 5.021 {\tiny(0.340)} & 5.076 {\tiny(0.813)}
& 63.2\% {\tiny(1000)} & 2.056 {\tiny(0.476)} & 6.494 {\tiny(0.657)} \\

Swahili
& 32.2\% {\tiny(665)}  & 7.873 {\tiny(0.552)} & 7.070 {\tiny(0.778)}
& 4.3\%  {\tiny(88)}   & 4.876 {\tiny(0.234)} & 4.048 {\tiny(1.723)}
& 63.6\% {\tiny(1313)} & 2.129 {\tiny(0.426)} & 7.151 {\tiny(0.649)} \\

\bottomrule
\end{tabular}
\end{adjustbox}
\caption{\textbf{VA statistics by language.} The data are partitioned into positive (V>5.5), neutral (4.5<=V<=5.5), and negative (V<4.5) subsets. For each subset, we report the proportion of stance instances (\%) and the corresponding counts (n), as well as the mean VA values with standard deviations (SD) shown in {\tiny tiny}.}
\label{tab:valence_arousal_language_pct}
\end{table*}

\section{Dataset Analysis}

This section provides an analysis of the VA distributions of stance
instances and target usage.  

\smallskip %
\noindent\textbf{{U-shaped}  distributions.} \Cref{fig:va_scatter} shows the VA distributions of stance instances across different languages and domains. %
The red shaded regions visualize the joint VA distribution of stance instances, with darker colors indicating higher instance density.
The black curve illustrates the relationship between valence and arousal.
Overall, all datasets exhibit a broadly U-like relationship between valence and arousal; that is, arousal tends to be lowest around neutral valence and increases toward both negative and positive extremes.

\begin{table*}[ht]
\centering
\small
\setlength{\tabcolsep}{3.5pt}
\renewcommand{\arraystretch}{1.15}

\newcommand{\TopTenFixedH}{0.13\textheight}

\begin{adjustbox}{totalheight=\TopTenFixedH, keepaspectratio}
\begin{tabular}{r|l r r|l r r|l r r}
\hline
\textbf{Rank} &
\textbf{English} (env. protection) & \textbf{n (\%)}& \textbf{Mean V\#A} &
\textbf{German} (politics) & \textbf{n (\%)}& \textbf{Mean V\#A} &
\textbf{Chinese} (env. protection) & \textbf{n (\%)}& \textbf{Mean V\#A} \\
\hline
--- &
\textit{Overall} & 4211 (100\%) & 5.18\#6.48 &
\textit{Overall} & 1848 (100\%) & 5.76\#5.76 &
\textit{Overall} & 2138 (100\%) & 5.02\#5.42 \\
\hline
1  & nuclear energy & 230 (5.46\%) & 6.37\#6.96 &
Bildung {\tiny (education)} & 19 (1.03\%) & 7.32\#6.41 &
環保 {\tiny (environmental protection)} & 35 (1.64\%) & 5.55\#5.53 \\
2  & climate change & 157 (3.73\%) & 3.12\#6.55 &
Schuldenbremse {\tiny (debt brake)} & 18 (0.97\%) & 3.99\#6.30 &
環境 {\tiny (environment)} & 19 (0.89\%) & 5.62\#5.41 \\
3  & EPA & 127 (3.02\%) & 4.96\#6.03 &
ÖPNV {\tiny (public transport)} & 15 (0.81\%) & 7.28\#6.29 &
核能 {\tiny (nuclear energy)} & 18 (0.84\%) & 5.17\#5.45 \\
4  & Electric vehicles & 112 (2.66\%) & 6.28\#6.55 &
Gesellschaft {\tiny (society)} & 13 (0.70\%) & 6.98\#5.80 &
地球 {\tiny (Earth)} & 17 (0.80\%) & 5.67\#5.41 \\
5  & clean energy & 101 (2.40\%) & 6.76\#7.10 &
Frauen {\tiny (women)} & 13 (0.70\%) & 5.72\#6.09 &
再生能源 {\tiny (renewable energy)} & 14 (0.65\%) & 4.95\#5.25 \\
6  & coal plant & 59 (1.40\%) & 3.40\#6.39 &
Wohnraum {\tiny (housing)} & 13 (0.70\%) & 6.05\#6.33 &
綠能 {\tiny (green energy)} & 14 (0.65\%) & 5.03\#5.43 \\
7  & Environmental awareness & 36 (0.85\%) & 7.02\#7.16 &
Landwirtschaft {\tiny (agriculture)} & 12 (0.65\%) & 6.34\#5.51 &
能源轉型 {\tiny (energy transition)} & 13 (0.61\%) & 5.03\#5.32 \\
8  & nuclear & 24 (0.57\%) & 7.27\#7.19 &
Polizei {\tiny (police)} & 11 (0.60\%) & 5.77\#5.14 &
風電 {\tiny (wind power)} & 13 (0.61\%) & 4.53\#5.35 \\
9  & government & 15 (0.36\%) & 3.83\#6.29 &
Verfassungsschutz {\tiny (domestic intelligence)} & 11 (0.60\%) & 4.65\#6.21 &
氣候變遷 {\tiny (climate change)} & 12 (0.56\%) & 4.08\#5.44 \\
10 & coal & 14 (0.33\%) & 2.98\#6.83 &
Infrastruktur {\tiny (infrastructure)} & 11 (0.60\%) & 6.65\#5.82 &
綠電 {\tiny (green electricity)} & 11 (0.51\%) & 4.44\#5.25 \\
\hline
\end{tabular}
\end{adjustbox}

\vspace{0.6em}

\begin{adjustbox}{totalheight=\TopTenFixedH, keepaspectratio}
\begin{tabular}{r|l r r|l r r}
\hline
\textbf{Rank} &
\textbf{Nigerian Pidgin} (politics) & \textbf{n (\%)}& \textbf{Mean V\#A} &
\textbf{Swahili} (politics) & \textbf{n (\%)}& \textbf{Mean V\#A} \\
\hline
--- &
\textit{Overall} & 1583 (100\%) & 3.89\#6.03 &
\textit{Overall} & 2066 (100\%) & 4.10\#6.99 \\
\hline
1  & peter obi & 114 (7.20\%) & 6.49\#5.67 & wewe {\tiny (you)} & 70 (3.39\%) & 2.74\#7.26 \\
2  & atiku & 65 (4.11\%)  & 3.99\#6.10 & mimi {\tiny (me)} & 38 (1.84\%) & 4.86\#7.17 \\
3  & inec & 63 (3.98\%)  & 1.85\#7.03 & pussy {\tiny (vulgar term)} & 27 (1.31\%) & 3.72\#7.02 \\
4  & you & 55 (3.47\%)  & 2.96\#6.56 & huyu {\tiny (this person)} & 27 (1.31\%) & 2.26\#7.20 \\
5  & @officialabat & 38 (2.40\%)  & 4.17\#6.02 & yeye {\tiny (he/she)} & 25 (1.21\%) & 3.34\#7.22 \\
6  & Tinubu & 32 (2.02\%)  & 3.30\#6.35 & wakikuyu {\tiny (Kikuyu people)} & 19 (0.92\%) & 2.75\#7.17 \\
7  & election & 29 (1.83\%)  & 3.16\#6.32 & wajaluo {\tiny (Luo people)} & 17 (0.82\%) & 2.38\#7.28 \\
8  & Kwankwaso & 26 (1.64\%)  & 6.32\#5.88 & mtombaji {\tiny (suitor)} & 16 (0.77\%) & 2.89\#7.11 \\
9  & apc & 24 (1.52\%)  & 2.45\#6.68 & matako {\tiny (buttocks)} & 11 (0.53\%) & 2.18\#7.01 \\
10 & Nigeria & 23 (1.45\%)  & 4.59\#5.56 & nigga {\tiny (racial slur)} & 10 (0.48\%) & 2.25\#6.97 \\
\hline
\end{tabular}
\end{adjustbox}

\caption{\textbf{Top 10 targets per language.} For each target, the count (n) and the corresponding percentage (\%) are reported, along with the mean VA values. The \textit{Overall} row reports the total number of targets and overall mean VA values. Similar targets are grouped (e.g., \textit{Environmental Protection Agency}
and \textit{EPA}, or \textit{coal plant} and \textit{coal plants}).}
\label{tab:top10aspects}
\end{table*}

\smallskip %
\noindent\textbf{\textbf{Target usage analysis}.} Despite the U-like structural consistency observed in VA distributions, their distribution and symmetry vary across languages and domains.
The English data exhibit a relatively balanced valence distribution, with around 50\% positive and 39\% negative instances (cf. \Cref{tab:valence_arousal_language_pct}).  The arousal distribution is concentrated in the high-value region across both valence polarities (positive mean: 6.845; negative mean: 6.476). This finding suggests that environmental stance expressions are typically associated with increased arousal. For example, as shown in \Cref{tab:top10aspects}, the negatively valenced expressions include concerns about the urgency of \textit{climate change}, opposition to \textit{coal} and \textit{coal plants}, as well as criticism of perceived \textit{government} inaction. On the positive side, these expressions include supportive orientations toward \textit{environmental awareness} and related energy-transition targets. Notably, although \textit{EPA} is associated with near-neutral valence overall, this pattern is largely driven by a mixture of positive and negative expressions. 

The German data show a higher concentration of positive valence (69\% positive vs 25\% negative), with arousal at a moderate to high level (overall mean: 5.76; \Cref{tab:top10aspects}).
This is likely due to the source data, which are written by political parties.
The most frequent targets, such as \textit{Bildung (education)} and \textit{ÖPNV} \textit{(public transport)}, are consistent with this observation, as these policy areas are typically proposed as improvements following an election. In contrast, \textit{Schuldenbremse }\textit{(debt brake)} emerges as the only top ten target associated with negative valence.

\begin{table*}[t]
\centering
\begin{adjustbox}{width=\textwidth}
\begin{tabular}{l|ccc|ccc|cccc}
\toprule
& \multicolumn{3}{c|}{\textbf{PLM regressors}}
& \multicolumn{3}{c|}{\textbf{Closed LLMs (Prompting)}}
& \multicolumn{4}{c}{\textbf{LLM regressors (Fine-tuning)}} \\
\cmidrule(lr){2-4}\cmidrule(lr){5-7}\cmidrule(lr){8-11}
\textbf{Language}
& \textbf{RemBERT} & \textbf{LaBS} & \textbf{XLM-R}
& \makecell{\textbf{GPT-5}\\\textbf{mini}}
& \makecell{\textbf{Gemini-2.5}\\\textbf{Flash}}
& \makecell{\textbf{Kimi}\\\textbf{K2}}
& \makecell{\textbf{Mistral-3}\\\textbf{14B}}
& \makecell{\textbf{Phi-4}\\\textbf{14B}}
& \makecell{\textbf{Qwen-2.5}\\\textbf{72B}}
& \makecell{\textbf{Llama-3.3}\\\textbf{70B}} \\
\midrule
English
& \textbf{1.993} & 2.288 & 2.005
& 1.819 & 1.771 & \textbf{1.643}
& 1.643 & 1.575 & 1.520 & \textbf{1.468} \\
Chinese
& 0.795 & \textbf{0.715} & 0.747
& 1.492 & 1.284 & \textbf{1.044}
& 0.740 & 0.691 & 0.681 & \textbf{0.679} \\
German
& 2.262 & 1.918 & \textbf{1.667}
& 1.754 & 1.701 & \textbf{1.671}
& 1.591 & 1.535 & 1.493 & \textbf{1.415} \\
Nigerian Pidgin
& 2.149 & \textbf{1.720} & 1.732
& 1.389 & 1.353 & \textbf{1.284}
& 1.739 & 1.409 & 1.228 & \textbf{1.157} \\
Swahili
& 2.322 & 2.103 & \textbf{2.054}
& 2.321 & 2.259 & \textbf{2.242}
& 2.299 & 2.055 & 2.012 & \textbf{1.859} \\

\midrule
AVG
& 1.904 & 1.749 & \textbf{1.641}
& 1.755 & 1.674 & \textbf{1.577}
& 1.602 & 1.453 & 1.387 & \textbf{1.316} \\
\bottomrule
\end{tabular}
\end{adjustbox}
\caption{\textbf{Comparison of PLM and LLM approaches.} RMSE of dimensional stance regression. Bold indicates the lowest (best) result within each model family for each language. Closed LLMs use 16-shot prompting.}
\label{tab:main_results}
\end{table*}

The Chinese data display a more compact distribution, with both positive and negative instances clustering around mid-range valence (positive mean: 6.068; negative mean: 3.893) and moderate arousal (overall mean: 5.42). The distribution shows limited dispersion along both dimensions, suggesting that stance expressions are generally mild.
Further analysis reveals substantial overlap in the top-10 most frequent targets between Chinese and English.
Notably, energy-transition–related targets in Chinese tend to have a closer neutral valence. This observation suggests that even when targeting similar environmental issues, stance expressions may exhibit different affective profiles across languages, potentially influenced by cultural, linguistic, and socio-contextual factors.

The Nigerian Pidgin data show a sparsely populated neutral region, resulting in limited continuity in the valence distribution. In addition, the distribution is dominated by negative valence (63\% negative vs. 36\% positive), and higher arousal is also observed at the negative valence end (mean: 6.494). This distribution can be attributed to the electoral nature of the source data. The most frequent targets also largely consist of political figures (e.g., \textit{Peter Obi}, \textit{Atiku}) and political institutions (e.g., \textit{INEC}, \textit{APC}), which are particularly likely to trigger emotionally charged political discussions and are often framed in a negative or critical manner in election-related discourse.   

Similar to Nigerian Pidgin, Swahili data also leans towards negative valence (64\% negative vs 32\% positive).
Upon inspection of the most frequent targets, we find that Swahili data contains a noticeable number of toxic examples (e.g., \textit{pussy}, \textit{nigga}).
Also, Swahili data shows the highest mean arousal (6.99) towards the targets.
Nigerian Pidgin and Swahili data originate from the X platform, written by citizens of the respective countries. 
By contrast, the German source data is weighted towards positive-valence examples and is produced by political parties.
Thus, the difference between political parties' and the general public's perception is evident when comparing these languages.

\section{Experiments}

This section describes the evaluation of the \datasetname datasets under the dimensional stance regression task, including the experimental setup, baseline results, and the corresponding analysis.

\begin{figure*}[tbp]
    \centering    \includegraphics[width=\linewidth]{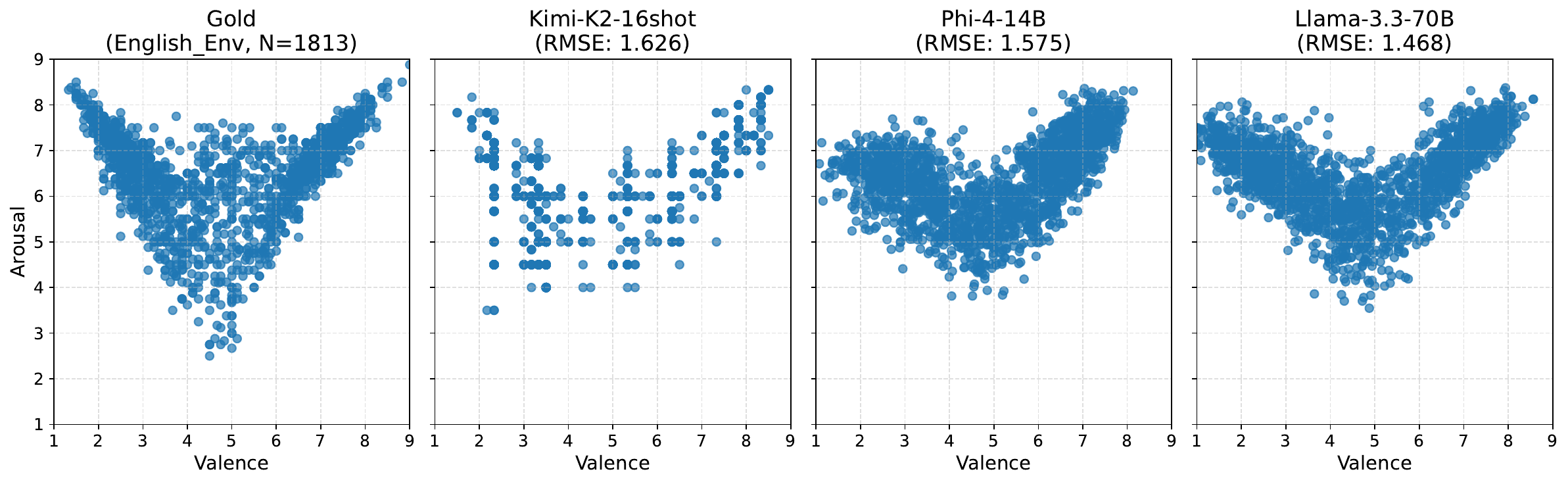}
    \caption{\textbf{English VA prediction scatter plots.} Comparison of gold and predicted VA distributions from selected models on the English test set. The x-axis shows valence, and the y-axis shows arousal on a one to nine scale.} 
    \label{fig:eng_EP_scatter}
\end{figure*}

\subsection{Setup}

\paragraph{Task.}
The goal of dimensional stance regression is to predict VA scores for each target aspect within a given text on a continuous scale from one to nine. The data split sizes are reported in \Cref{tab:final_data_stats}.  

\paragraph{{Models.}}
We evaluate various PLMs and LLMs across three groups.

\smallskip %
\noindent\textbf{PLM regressors.}
This group provides a baseline under numerical regression, representing prevailing approaches to structured text regression using traditional encoder-based models. We employ three multilingual PLMs, XLM-R \cite{conneau2020unsupervised}, RemBERT \cite{chung2021rethinking}, and LaBSE \cite{feng2022language}, each trained independently on the training split of its respective dataset. For these models, a dedicated regression head comprising a linear layer and dropout is applied to the sentence-level representation. The entire architecture is optimized end-to-end via mean squared error (MSE) loss to predict VA scores. 

\smallskip %
\noindent\textbf{Closed LLMs (prompting).}
This group provides a baseline under free-form numerical generation, leveraging advanced LLMs through prompting without parameter updates. Three closed-source LLMs, GPT-5 mini \cite{OpenAIGPT-52025}, Gemini 2.5 Flash \cite{gemini2025report}, and Kimi K2 \cite{moonshot2025kimik2}, are evaluated under zero-shot and few-shot settings. For the few-shot setting, in-context examples are selected from the first \textit{$k$} samples in the training split. These models are accessed via their respective APIs to directly generate VA scores based on their inherent reasoning. 

\smallskip %
\noindent\textbf{LLM regressors (fine-tuning).}
We provide a numerical regression baseline by adapting LLM representations for VA prediction. Four open-source LLMs, \texttt{Ministral-3-14B} \cite{mistral3_2025}, \texttt{Phi-4-14B} \cite{phi4_2024}, \texttt{Llama-3.3-70B} \cite{llama3_2024}, and \texttt{Qwen-2.5-72B} \cite{qwen25_2024}, are adapted via 4-bit QLoRA \cite{dettmers2023qlora} for parameter-efficient fine-tuning. Following the PLM setup, a regression head is attached to the hidden state of the last token. The LoRA adapters and the regression layer are jointly optimized using MSE to estimate VA scores. Further implementation details are given in \Cref{app:implementation}.

\paragraph{Evaluation metric.}
We adopt RMSE to measure prediction error in the VA space, defined as 
\begin{equation}
\small
RMSE_{VA} = \sqrt{\sum_{i=1}^N 
   \frac{(V_p^{} - V_g^{})^2 + (A_p^{} - A_g^{})^2}{N} }
\end{equation}  
where $N$ is the total number of instances; $V_p^{}$ and $A_p^{}$ denote the predicted valence and arousal values for an instance; and $V_g^{}$ and $A_g^{}$ denote the corresponding gold values.  

\subsection{Experimental Results} 

\Cref{tab:main_results} summarizes the results for dimensional stance regression across PLMs and LLMs. PLM regressors provide baseline results under numerical regression. Within the PLM family, different models exhibit language-specific strengths, with no single model dominating across all languages. Overall, XLM-R achieves the lowest average RMSE across languages. 

Closed LLMs are evaluated under generative numerical prediction using 16-shot prompting. Among these models, Kimi K2 achieves the strongest overall performance across languages. When compared with PLM regressors trained on the full training set, LLMs with few-shot prompting achieve competitive or better performance on average, particularly in English, German, and Nigerian Pidgin, while underperforming in Chinese and Swahili. Nevertheless, prompting-based LLMs can serve as data-efficient baselines for dimensional stance regression.  

For LLM regressors, fine-tuned models generally outperform prompted LLMs, yielding lower RMSE in most languages. More consistent gains are observed for larger 70B-scale models. These findings indicate that, beyond prompting, LLM representations can be effectively adapted for regression, offering a viable alternative for modeling continuous VA scores.  

\begin{figure}[t]
    \centering    \includegraphics[width=\linewidth]{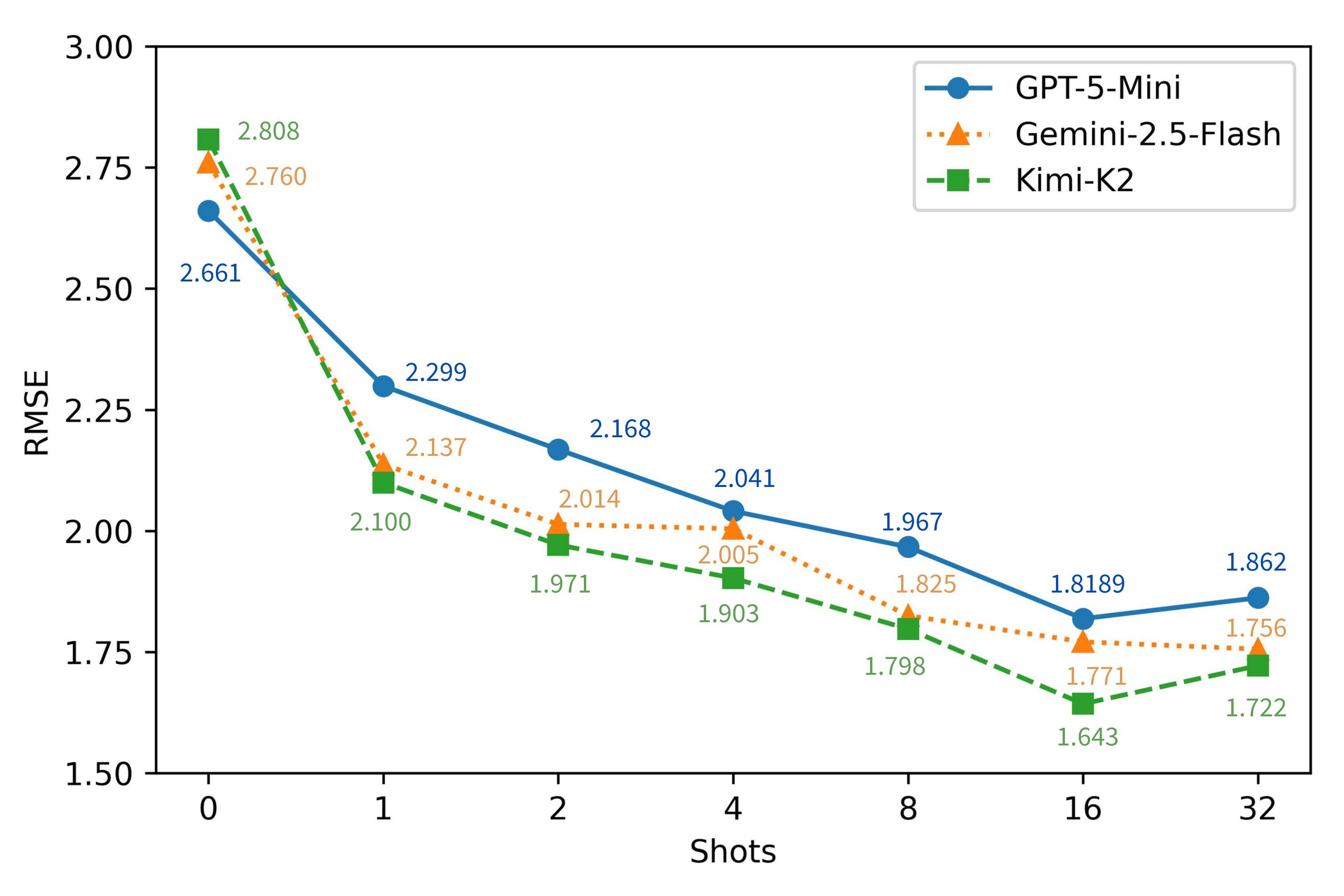}
    \caption{\textbf{Trendlines of few-shot performance.} We compare the prompting of three closed LLMs. The x-axis represents the number of shots, ranging from zero to a 32-shot setup. The y-axis shows the RMSE.} 
    \label{fig:Fewshot_trendline}
\end{figure}

Examining the results by language, Chinese consistently exhibits the best performance across models. This can be attributed to its concentrated VA distribution around the neutral region, with a limited range of variation. In contrast, performance on Swahili is generally lower, likely reflecting the challenges of low-resource languages. Nigerian Pidgin does not show a comparable decline, possibly due to its English-like characteristics.

\subsection{Analysis}

\begin{table}[tbp]
    \small
    \setlength{\tabcolsep}{3.5pt}
    \renewcommand{\arraystretch}{1.15}
    \begin{adjustbox}{width=\columnwidth}
    \begin{tabular}{lccc}
    \toprule
    \textbf{Model} & \textbf{Prompt v1} & \textbf{Prompt v2} & \textbf{Prompt v3} \\
    \midrule
    \textbf{GPT-5 mini}& 2.661 & 2.717 & \textbf{2.406} \\
    \textbf{Gemini-2.5 Flash}& 2.760 & 3.118 & \textbf{2.651} \\
    \textbf{Kimi K2}& 2.808 & 2.413 & \textbf{2.073} \\
    \bottomrule
    \end{tabular}
    \end{adjustbox}
    \caption{\textbf{Sensitivity to prompt phrasing.} Performance of LLMs across three prompt paraphrases on the English test set. The prompts are shown in \Cref{app:prompts}.}
    \label{tab:prompt_ablation}
\end{table}

\paragraph{\textbf{Alignment of predicted and gold VA distributions.}}
 \Cref{fig:eng_EP_scatter} presents the gold and predicted VA distributions on the English test set \footnote{Figures for other languages are provided in \Cref{app:pred_comparisons}.}. LLM prompting produces grid-like (coarse-score) VA distributions across all languages, likely due to its token-based generation, which tends to yield discrete numerical outputs when predicting continuous VA scores. In contrast, distributions produced by LLM regressors are more continuous, aligning closely with the gold distributions, particularly for the 70B model. The disadvantages of discretized outputs from LLM prompting become particularly evident under compact VA distributions with limited variation, such as Chinese, as reflected by their higher RMSE compared to all regressors. In such cases, regression-based models can more easily align with the gold distributions. 

\paragraph{\textbf{Effect of few-shot examples.}}
\Cref{fig:Fewshot_trendline} shows that increasing the number of shots on the English test set consistently improves performance across models. Performance tends to reach a plateau at 16 shots for all closed LLMs except for Gemini 2.5 Flash. Notably, Kimi K2 ultimately outperforms the other models despite exhibiting the lowest performance in the zero-shot setting.

\paragraph{\textbf{Effect of prompt variations.}}
We test the paraphrased variants of prompts outlined in \Cref{app:prompts}.
As shown in \Cref{tab:prompt_ablation}, the performance of closed LLMs appears to be highly dependent on prompt wording when predicting VA scores on the English test set using different paraphrases of the same instruction.

\section{Related Work}

\textbf{Categorical stance.} Prior work on categorical stance detection has expanded primarily along three axes: language coverage, scale, and domain specificity. Early benchmarks focused on English Twitter data with coarse labels (favor/against/neutral), such as the SemEval stance dataset of \citet{MohammadKSZ16a}. Multilingual extensions followed, including IberEval 2017–2018 for Catalan and Spanish tweets on the Catalan independence referendum \citep{ZotovaANR20}, and X-Stance, which covers German, French, and Italian political discourse using question-based target representations \citep{VamvasS20b}. Large-scale English resources include P-Stance \citep{LiSSN21} and COVID-19-Stance \citep{GlandtKLC21}, both centered on prominent political or societal targets. More recent benchmarks emphasize generalization and zero-shot settings, such as C-STANCE for Chinese \citep{zhao-etal-2023-c} and EZ-STANCE for English with claim- and noun-phrase targets \citep{zhao-caragea-2024-ez}. PolitiSky24 introduces user-level and platform-specific perspectives on Bluesky \citep{rostami2025politisky24uspoliticalbluesky}. Domain-focused datasets such as ClimateStance and ClimateEng \citep{VaidPS22} broaden topical coverage but continue to rely on categorical stance labels.

\noindent\textbf{Emotion-aware stance.} While categorical stance remains an established field, the emergence of dimensional sentiment and emotion analysis \cite{mohammad2018obtaining, lee2022chinese,lee2024overview,muhammad-etal-2025-brighter} has led to recent work that investigates how continuous emotion features can aid stance analysis.  
PoliStance‑VAE disentangles latent valence, arousal, and dominance using a variational autoencoder and achieves state‑of‑the‑art cross‑target results on P‑Stance and SemEval‑2016, showing the utility of fine‑grained emotional representations \citep{XuLA25}.
TWISTED combines toxicity, morality, and speech‑act cues with lexicon‑derived valence, arousal, and dominance to improve stance detection across datasets such as SemEval‑2016, P‑Stance, COVID‑19‑Stance, and climate change \citep{UpadhyayaFN23a}.
For climate change tweets, adding emotion recognition and intensity prediction helps distinguish climate change deniers, whose posts often contain high-intensity anger and disgust, from believers, whose tweets express anticipation and trust \citep{UpadhyayaFN23}.
ClimaConvo is a large Twitter corpus covering relevance, stance, hate speech, humor, and related dimensions, enabling deeper analysis of climate change discourse \citep{shiwakoti-etal-2024-analyzing}.
Other work analyzes how rich representations of stance contexts across Reddit communities reveal patterns of community identity and engagement beyond lexical similarity \citep{aggarwal-etal-2023-investigating}.

\noindent\textbf{Research gap.} The above emotion-aware stance studies primarily rely on dimensional sentiment resources to enhance stance analysis. However, a critical gap remains: the lack of human-annotated dimensional datasets for stance analysis. This limitation is particularly pronounced for low-resource languages such as Swahili and Nigerian Pidgin, which currently lack even basic stance resources.

\section{Conclusion}

We present \datasetname, the first dimensional stance resource with human VA annotations, covering five high- and low-resource languages across two domains. 
Building on this resource, we introduce dimensional stance regression as a unified framework for modeling stance through affective representations. Our analysis shows that the VA space effectively captures the diverse ways stances are expressed across languages and domains. This complements traditional polarity-based analysis with a more nuanced and affective perspective.
Experiments with regression and numerical generation baselines reveal substantial performance gaps in low-resource languages and highlight the limitations of token-based VA prediction, underscoring key challenges and opportunities for advancing multilingual fine-grained stance analysis.

\section*{Limitations}

Although \datasetname is multilingual, interpretations of valence and arousal can vary across cultures, thereby affecting cross-lingual comparability.
We mitigate this by using five native-speaker annotators per language and sample, consistent 1–9 VA scales, and shared guidelines; nonetheless, results should be interpreted as comparisons across language-community-domain settings. Expanding language coverage and testing measurement invariance are important directions for future work.

Inter-annotator agreement is acceptable, yet consistently lower for arousal than for valence (e.g., 1.70 vs 2.20 RMSE for English).
This observation is consistent with previous studies reporting lower reliability for arousal annotations \cite{buechel-hahn-2017-emobank, mohammad2018obtaining, lee2022chinese}.
We average ratings across five annotators and provide explicit guidance.
Downstream models should account for this uncertainty (e.g., confidence intervals, robustness checks) \citep{troiano-etal-2021-emotion}.

Some datasets (e.g., Nigerian Pidgin) include more samples of negative or positive valence, which may bias models during training and inflate performance in the majority regions of the VA space.
We document these differences and encourage explicit handling (e.g., reweighting or stratified sampling) when training or comparing models across languages \citep{10.5555/1622407.1622416}.

\section*{Ethical Considerations}

Stance is often conveyed in subtle, nuanced, and complex ways that may not even be the core intention of the speaker's utterance. Furthermore, there can be considerable variation in how individuals express their stance. It should be noted that human-annotated labels of stance capture \textbf{perceived} stance, and that in several cases this may be different from the speaker's true stance. Nonetheless, since language is a key mechanism to communicate, at an aggregate level, perceived stance tends to correlate with true stance. Thus, even a perceived stance is useful at an aggregate level. However, caution must be exercised when using individual inferred stances to make decisions about individuals, especially those involving high-stakes consequences.  

Similar to other enabling technologies, stance detection can be used in inappropriate and harmful ways to society. For example, it can be used to identify individuals who are opposed to certain government policies, and this information can be used to unfairly persecute those individuals. Another troubling, and perhaps more pernicious example, is the use of stance detection to manipulate user behavior. For example, voting in a certain way or purchasing products. For example, if it is determined that one cares deeply about a particular issue, that person can be bombarded with carefully doctored information to make them more politically polarized. 

We strictly prohibit the use of our data for commercial or government purposes. In many ways, the ethical considerations associated with stance detection overlap with those of automatic sentiment and emotion detection.
Thus, we also refer the interested reader to \citet{mohammad-2022-ethics-sheet} for a wide-ranging discussion of relevant ethical considerations.

\section*{Acknowledgements}

Jonas Becker acknowledges the support of the Landeskriminalamt NRW.
\newline
Jan Philip Wahle, Terry Ruas, and Bela Gipp acknowledge the support of the Lower Saxony Ministry of Science and Culture, and the VW Foundation.
\newline
Shamsuddeen Hassan Muhammad acknowledges the support of Google DeepMind, whose funding made this work possible.
\newline
Liang-Chih Yu and Lung-Hao Lee acknowledge the support of the National Science and Technology Council, Taiwan, under the grants NSTC 113-2221-E-155-046-MY3 and NSTC 114-2221-E-A49-059-MY3.

\bibliography{custom, dimStance}

\clearpage

\appendix

\section*{Appendix}
\label{sec:appendix}

\section{Implementation details.}\label{app:implementation}
All experiments are implemented using the PyTorch-based implementations of both PLMs and LLMs provided by the Hugging Face Transformers library. We adopt the AdamW optimizer with a linear learning rate scheduler. The learning rate is fixed at 2e-5. The batch size is set to 16 for PLMs and 8 for LLMs. All models are fine-tuned for 5 training epochs. Experiments are conducted on NVIDIA H200 GPUs.

\section{Models Used}\label{app:models}
\subsection{PLMs}
 \begin{enumerate}
    \item \url{https://huggingface.co/google/rembert}
    \item \url{https://huggingface.co/sentence-transformers/LaBSE}
    \item \url{https://huggingface.co/FacebookAI/xlm-roberta-large}
\end{enumerate}

\subsection{Open-Source LLMs}
\begin{enumerate}
    \item \url{https://huggingface.co/mistralai/Ministral-3-14B-Reasoning-2512}
    \item \url{https://huggingface.co/microsoft/phi-4}
    \item \url{https://huggingface.co/Qwen/Qwen2.5-72B-Instruct}
    \item \url{https://huggingface.co/meta-llama/Meta-Llama-3.3-70B}
\end{enumerate}

\section{Annotation Guidelines} \label{app:annotation_guidelines}

We summarize the annotation guidelines below.
The dataset is annotated at the level of aspects extracted from stance texts. For each identified aspect, annotators assign affective ratings and assess aspect validity for automatically extracted aspects. The guidelines are designed to be consistent across languages, while allowing for language-specific realizations of aspects.

\paragraph{Annotation Units.}
A \emph{target aspect} denotes a concrete entity, attribute, or concept being discussed. A single text sample may contain multiple aspects, each annotated individually.
\begin{itemize}
    \item \textbf{Aspect}, a specific part, feature, or attribute of an entity (e.g., product, service, or topic) that is the target of an expressed opinion.
\end{itemize}

\paragraph{Affective Dimensions.}
Each valid aspect is annotated along two continuous emotional dimensions:
\begin{itemize}
    \item \textbf{Valence}, capturing the degree of positivity or negativity of the expressed opinion, ranging from high negative (1) through neutral (5) to high positive (9).
    \item \textbf{Arousal}, capturing the intensity or emotional activation of the opinion, ranging from highly unaroused (1) to highly aroused (9).
\end{itemize}
Annotators apply these dimensions jointly, following established affective patterns: emotionally charged opinions (both positive and negative) typically exhibit higher arousal, whereas neutral opinions tend to be associated with lower arousal. Implausible combinations of valence and arousal are avoided.

\paragraph{Validity Judgment.}
In addition to affective ratings, annotators determine whether an extracted aspect is \emph{valid}. Validity guidelines require:
\begin{itemize}
    \item splitting multiple aspects into separate aspects when applicable,
    \item inclusion of relevant modifiers such as negation or intensifiers within the expression,
    \item precise and minimal spans for both aspect terms.
\end{itemize}
Aspects that do not meet these criteria are marked as invalid and excluded from downstream use.

\paragraph{Cross-Lingual Consistency.}
Although the surface realization of aspects may vary across languages, the underlying annotation principles---including the definition of annotation units, affective dimensions, rating scales, and validity criteria---are applied uniformly across all language-specific datasets.

\onecolumn

\section{Dataset Examples} \label{app:dataset_examples}

\begin{table*}[h]
\centering
\small
\setlength{\tabcolsep}{6pt}
\renewcommand{\arraystretch}{1.25}
\begin{tabular}{l p{0.7\textwidth} r}
\hline
\textbf{Ex.} & \textbf{Textual instance} & \textbf{V\#A} \\
\hline
\textbf{1} &
It's great that after coal plants shut down there are transmission lines to connect \hl{solar and wind resources} too. Pennsylvania needs cleaner air.&
$7.75\#7.62$\\
\textbf{2} & ...\hl{@peterobi} your mandate can only be delayed but cannot be denied please my fellow obedient be calm.  
{\textit{\hl{@peterobi} your mandate can only be delayed but cannot be denied please my fellow supporters be calm}}& $6.20\#4.23$\\
\textbf{3} &
\ldots\ Der \hl{Verfassungsschutz} ist keine Lösung, sondern gehört aufgelöst.  
{\textit{The \hl{Office for the Protection of the Constitution} is not a solution, but should be dissolved.}}&
$2.10\#7.20$ \\
\textbf{4} & 我覺得政府在\colorbox{yellow}{綠能運輸的支持}上還不夠.
{\textit{I feel that the government's \hl{support for green transportation} is still insufficient.}}&
$4.25\#4.62$\\
\textbf{5} &
sisi piahakika \hl{nyie} ndio mlikua wapinzani haswa tunashukuru pia tunategemea kuonana siku za usoni zaidi na zaidi.
{\textit{We are certain \hl{you} were the real rivals, but we are thankful and looking to meet again more often in the future.}}& $4.80\#5.50$\\
\hline
\end{tabular}
\caption{Examples from the DimStance datasets. Valence\#arousal (V\#A) refers to the highlighted target.}
\label{tab:examples}
\end{table*}

\clearpage
\section{Annotation Interface} \label{app:annotation_interface}

\begin{figure}[h]
    \centering

    \begin{subfigure}[h]{0.9\textwidth}
        \centering
        \includegraphics[width=\textwidth]{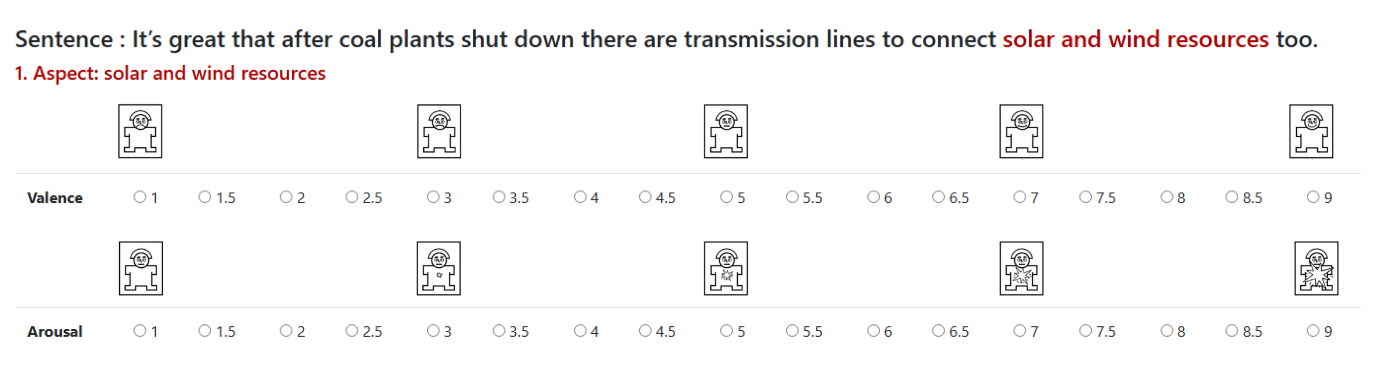}
        \caption{Annotation interface (English). Annotators label valence and arousal on a Likert scale from 1--9.}
        \label{fig:labelstudio_example_english}
    \end{subfigure}

    \vspace{0.8em}

    \begin{subfigure}[h]{0.9\textwidth}
        \centering
        \includegraphics[width=\textwidth]{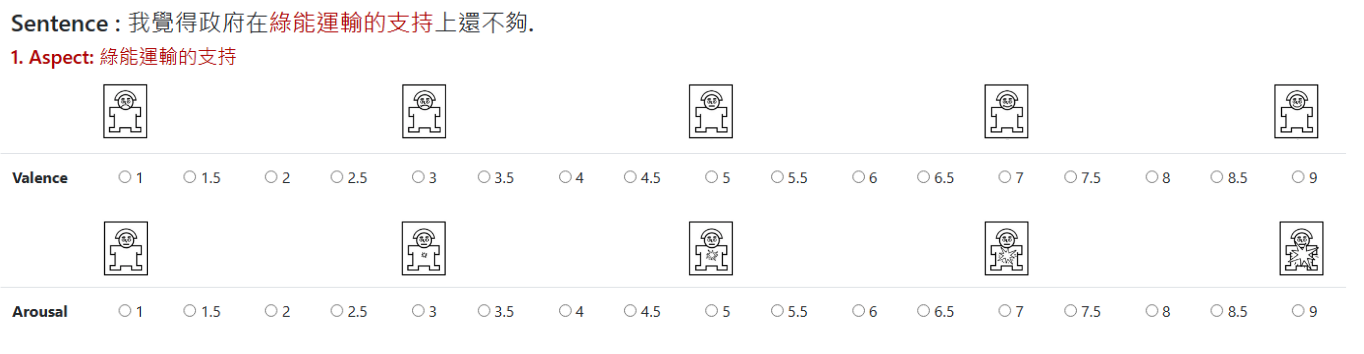}
        \caption{Annotation interface (Chinese). Annotators label valence and arousal on a Likert scale from 1--9.}
        \label{fig:labelstudio_example_chinese}
    \end{subfigure}

    \vspace{0.8em}

    \begin{subfigure}[h]{0.9\textwidth}
    \includegraphics[width=\textwidth]{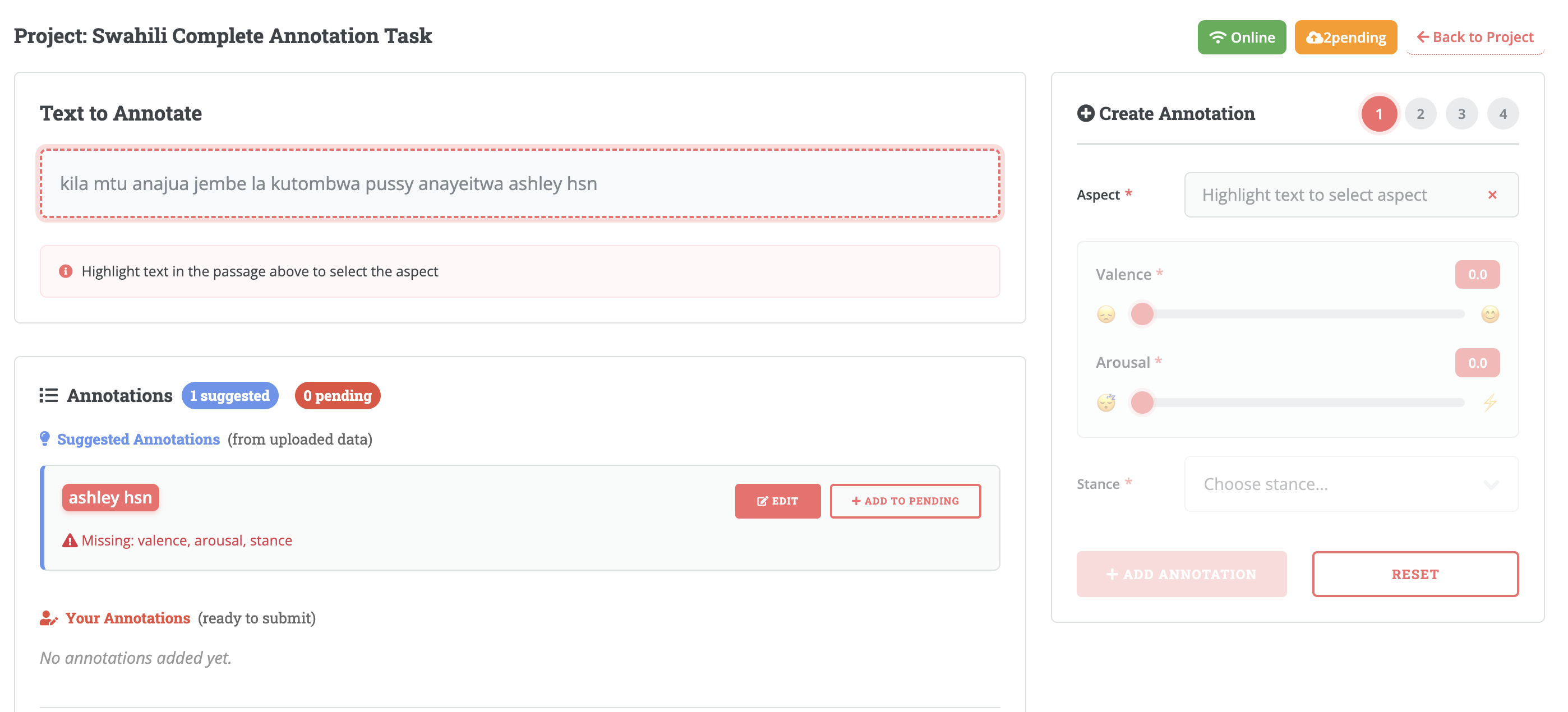}
    \caption{HausaNLP ABSA annotation tool (Nigerian Pidgin and Swahili). Annotators are presented with suggested aspects and provide valence and arousal scores on a 1--9 slider. Suggested aspects can be corrected, discarded, or extended.}
    \end{subfigure}
    \caption{Annotation interfaces used in our study.}
    \label{fig:annotation_interfaces}
\end{figure}

\clearpage
\section{Prediction Comparison}\label{app:pred_comparisons}

\begin{figure}[!htbp]
     \centering
     
     \begin{subfigure}[b]{0.8\linewidth}
         \includegraphics[width=\linewidth]{latex/figures/01-eng_EP_scatter.pdf}
         \caption{English}
     \end{subfigure}
     \par 
     
     \begin{subfigure}[b]{0.8\linewidth}
         \includegraphics[width=\linewidth]{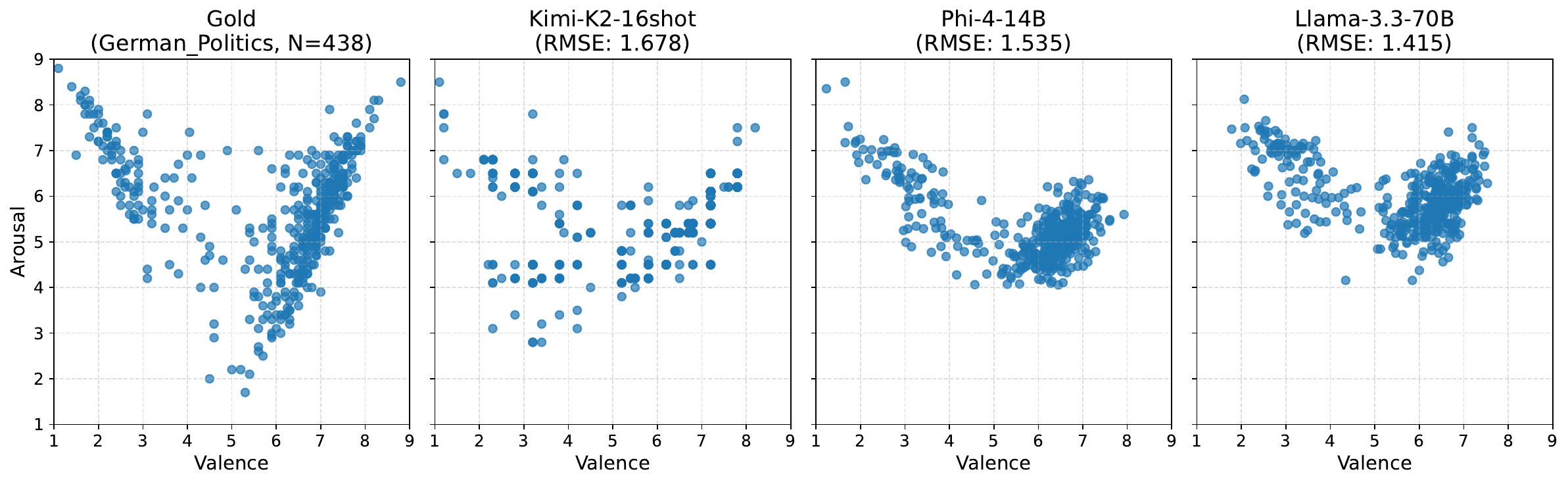}
         \caption{German}
     \end{subfigure}
     \par 

     \begin{subfigure}[b]{0.8\linewidth}
         \includegraphics[width=\linewidth]{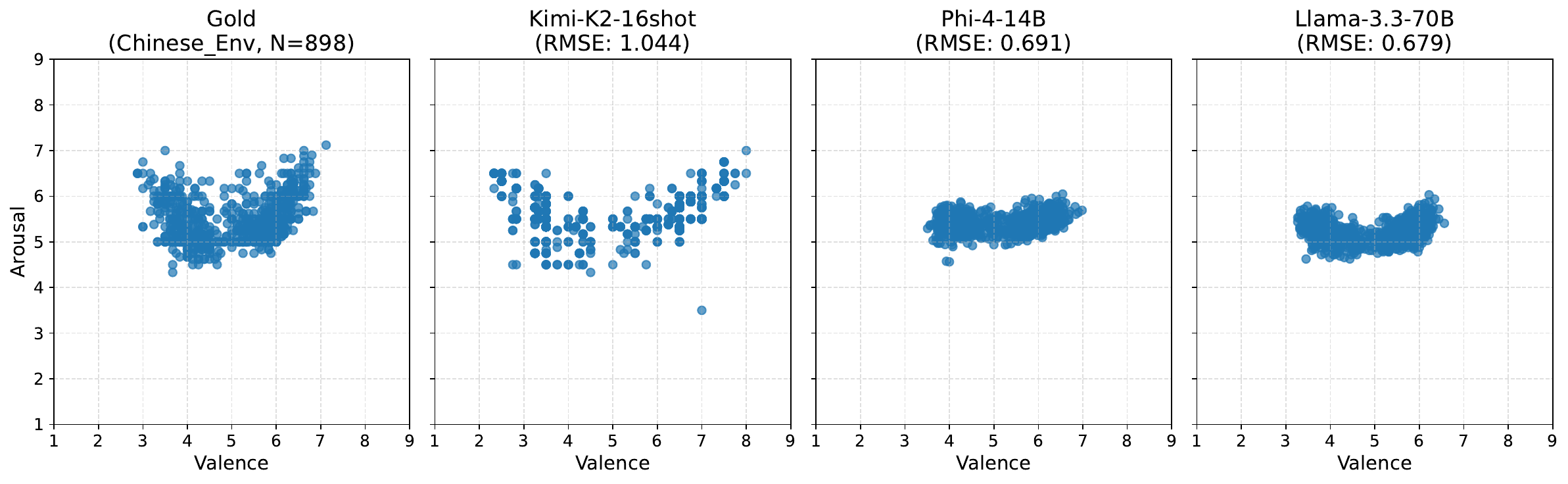}
         \caption{Chinese}
     \end{subfigure}
     \par 
     
     \begin{subfigure}[b]{0.8\linewidth}
         \includegraphics[width=\linewidth]{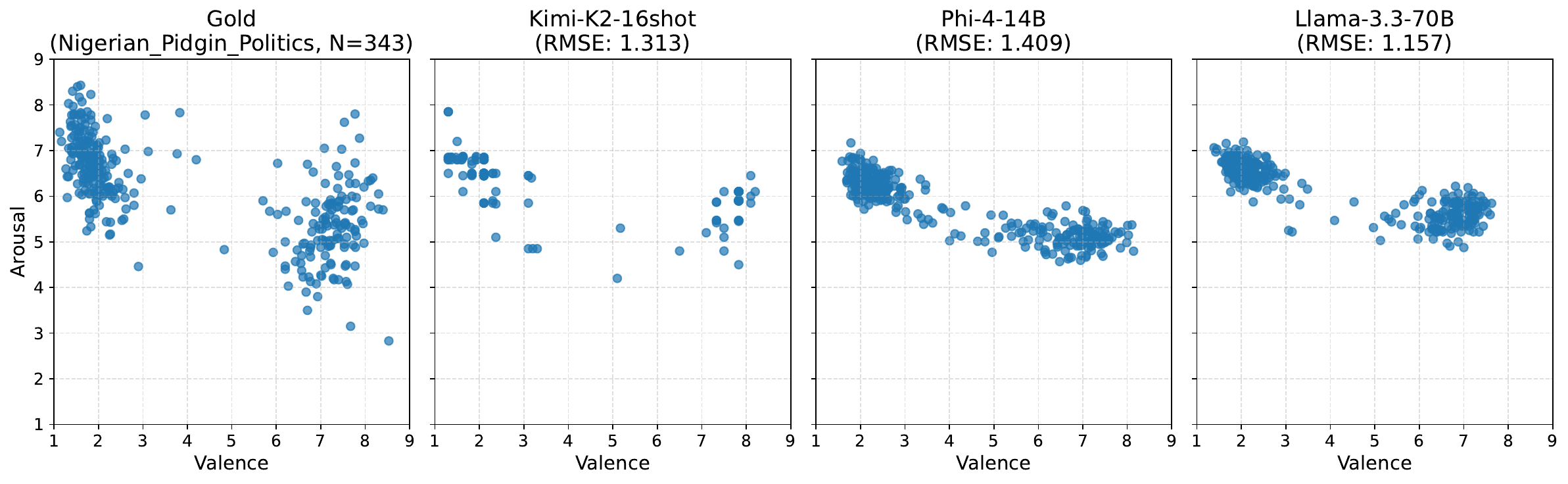}
         \caption{Nigerian Pidgin}
     \end{subfigure}
     \par 
     
     \begin{subfigure}[b]{0.8\linewidth}
         \includegraphics[width=\linewidth]{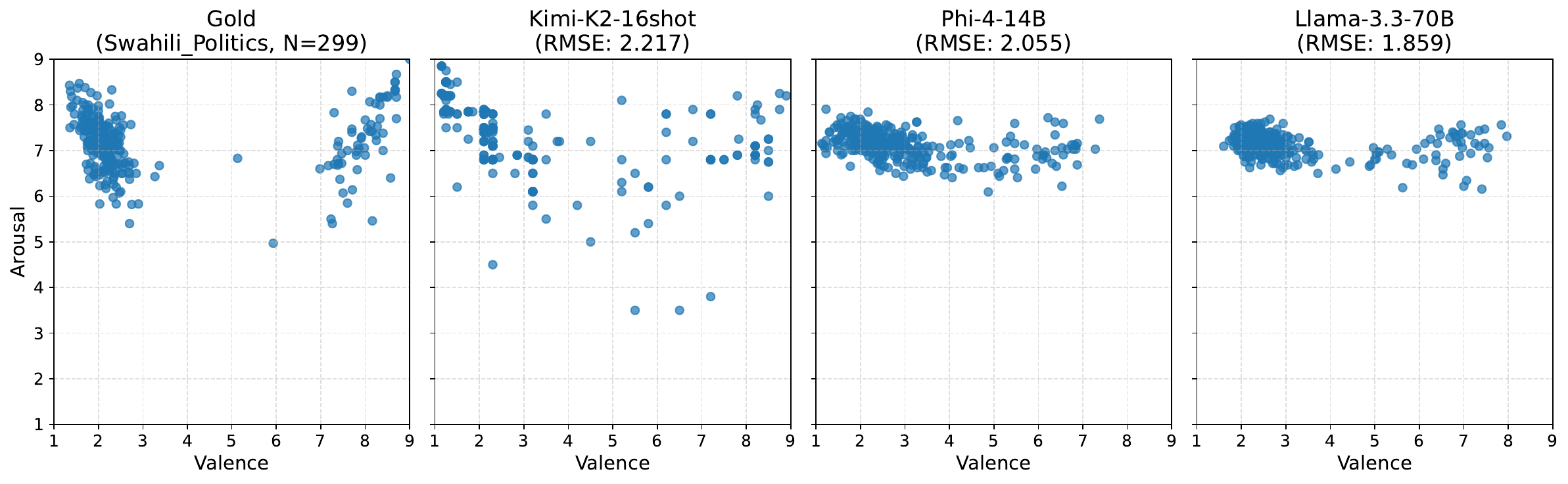}
         \caption{Swahili}
     \end{subfigure}
     
     \caption{Comparison between gold and predicted VA distributions across all test sets. }
     \label{fig:VA_scatter_all}
\end{figure}

\clearpage
\section{Prompts} \label{app:prompts}
\begin{table*}[h]
\centering
\begin{tabular}{p{3cm} p{10.5cm}}
\toprule
\textbf{Prompt Version} & \textbf{Prompt Text} \\
\midrule
Prompt v1 & 
    Definition: The output will be a pair of real numbers
    [valence, arousal] representing the sentiment toward the
    identified aspect in the sentence, where both values are on a
    scale from 1 to 9. 
    
    Valence: 1 = most negative, 9 = most 
    positive; 
    
    Arousal: 1 = calm/low intensity, 9 = excited/high 
    intensity. 
    
    Now complete the following example 
    
    <Examples> 
    
    \textbf{Input}: <sentence> The aspect is <aspect>. 
    
    \textbf{Output}: 
\\\\
Prompt v2 &
    Rate the sentiment toward the given aspect on two dimensions: 
    
     - Valence (1=very negative, 9=very positive) 
    
    - Arousal (1=calm, 9=excited) 
    
    <Examples> 
    
    Reply ONLY with two numbers separated by \#, e.g. 5.5\#6.2 
    
    \textbf{Text}: <sentence> Aspect: <aspect> 
        
    \textbf{Rating}:

\\\\
Prompt v3 &
    You are an expert sentiment analyst. Given a sentence and a     specific aspect mentioned in it, you must rate the expressed
    sentiment toward that aspect using two continuous scales:
    
    1.VALENCE (emotional positivity/negativity):
    1.0 = extremely negative, hostile, hateful
    5.0 = neutral, balanced, objective
     9.0 = extremely positive, enthusiastic, supportive 
     
    2.AROUSAL (emotional intensity/activation):
    1.0 = very calm, passive, subdued
    5.0 = moderate intensity
    9.0 = very intense, agitated, highly activated 
    
    Output format: <valence>\#<arousal> (e.g., 3.5\#7.2) 
    
    <Examples> 
    
    \textbf{Sentence}:<sentence>
    
    \textbf{Target aspect}: <aspect> 
    
    \textbf{Your rating}:

\end{tabular}
\caption{Prompt variants used in the ablation study for stance VA prediction.}
\label{tab:prompt-variants}
\end{table*}

\section{Usage of AI}
In the conduct of this research project, we used specific artificial intelligence tools and algorithms ChatGPT, Grammarly to assist with Writing. While these tools have augmented our capabilities and contributed to our findings, it's pertinent to note that they have inherent limitations. We have made every effort to use AI in a transparent and responsible manner. Any conclusions drawn are a result of combined human and machine insights.
This is an automatic report generated with AI Usage Cards \citep{aiusagecards}.

\end{document}